\documentclass[twoside,11pt]{article}

\usepackage{jmlr2e}

\usepackage[utf8]{inputenc} 
\usepackage[T1]{fontenc}    
\usepackage{hyperref}       
\usepackage{url}            
\usepackage{booktabs}       
\usepackage{amsfonts}       
\usepackage{nicefrac}       
\usepackage{microtype}      

\usepackage{bm}             
\usepackage{algorithm}      
\usepackage[noend]{algpseudocode} 
\usepackage{array}          
\usepackage{booktabs}       
\usepackage{pbox}           
\usepackage{subcaption} 
\usepackage{amsmath}
\usepackage{multirow}
\usepackage{wrapfig}

\newcommand{\minisection}[1]{\vspace{2mm}\noindent{\textbf{#1.}}}

\def\pNMLSingle{\underset{\theta \in \Theta}{\textit{max }} p_\theta (y|z^{N}, x, y)}
\newcommand\figref{Figure~\ref}
\newcommand\tableref{Table~\ref}
\newcommand\myalgref{Algorithm~\ref}

\newcommand{\ignore}[1]{}

\ShortHeadings{Deep pNML}{Bibas, Fogel and Feder}
\firstpageno{1}

\begin{document}

\title{Deep pNML: Predictive Normalized Maximum Likelihood for Deep Neural Networks}

%

\author{\name Koby Bibas \email kobybibas@gmail.com\\
        \addr School of Electrical Engineering \\
        Tel Aviv University 
        \AND
        \name Yaniv Fogel \email yaniv.fogel8@gmail.com \\
        \addr School of Electrical Engineering\\
        Tel Aviv University
        \AND
        \name Meir Feder \email meir@eng.tau.ac.il\\
        \addr School of Electrical Engineering\\
        Tel Aviv University}

\editor{}

\maketitle

\begin{abstract}
The Predictive Normalized Maximum Likelihood (pNML) scheme has been recently suggested for universal learning in the individual setting, where both the training and test samples are individual data. 
The goal of universal learning is to compete with a ``genie'' or reference learner that knows the data values, but is restricted to use a learner from a given model class. 
The pNML minimizes the associated regret for any possible value of the unknown label.
Furthermore, its min-max regret can serve as a pointwise measure of learnability for the specific training and data sample.
In this work we examine the pNML and its associated learnability measure for the Deep Neural Network (DNN) model class.
As shown, the pNML outperforms the commonly used Empirical Risk Minimization (ERM) approach and provides robustness against adversarial attacks. 
Together with its learnability measure it can detect out of distribution test examples, be tolerant to noisy labels and serve as a confidence measure for the ERM.
Finally, we extend the pNML to a ``twice universal'' solution, that provides
universality for model class selection and generates a learner competing with the best one from all model classes.
\end{abstract}


\section{Introduction} \label{sec:Introduction}

In the common situation of supervised machine learning, a trainset consisting of $N$ pairs of examples is given, $z^N=\{(x_i, y_i)\}_{i=1}^{N}$, where $x \in {\cal X}$ is the data or the feature and $y \in {\cal Y}$ is the label. Then, a new $x$ is given and the task is to predict its corresponding label $y$. 
The formal definition of the learning problem includes a loss function that measures the accuracy of the prediction. 
Here we assume that the learner assigns a probability $q(\cdot|x)$ to the test label, and we use the log-loss to evaluate the performance of the predictor
\begin{equation} \label{eq:logloss}
\ell(q;x,y) = -\log {q(y|x)}.
\end{equation}
Clearly, a reasonable goal is to find the predictor $q(\cdot|x)$ with the minimal loss. However, this problem is ill-posed unless additional assumptions are made.

First, a ``model'' class, or `hypotheses'' class must be defined. This class is a set of conditional probability distributions
\begin{align} 
P_\Theta = \{ p_\theta(y|x),\;\;\theta\in\Theta\} 
\end{align} 
where $\Theta$ is a general index set. 
This is equivalent to saying that there is a set of stochastic functions  $\{ y=g_\theta(x),\;\;\theta\in\Theta\}$ used to explain the relation between $x$ and $y$.

Next, assumptions must be made on how the data and the labels are generated. 
The most common setting in learning theory is
Probably Approximately Correct (PAC), established in \cite{valiant1984theory} where
$x$ and $y$ are assumed to be generated by some source $P(x,y)=P(x)P(y|x)$. $P(y|x)$ is not necessarily a member of $P_\Theta$. The goal is to perform as well as a learner that knows the best member of $P_\Theta$.

Another possible setting for the learning problem, recently suggested in \cite{Fogel2018} following, e.g., \cite{universal_prediction}, is
the individual setting, where the data and labels of both the training and test are specific and individual values.
In this setting the goal is to compete with a ``genie'' or a reference learner that knows the desired label value, but is restricted to use a model from the given hypotheses class $P_\Theta$, and that does not know which of the samples is the test. This learner then chooses:
\begin{align} \label{eq:genie} 
\hat{\theta}(z^N,x,y)  = \arg\max_\theta \left[ p_\theta(y|x) \cdot\Pi_{i=1}^N p_\theta(y_i|x_i) \right] .
\end{align}
The log-loss difference between a universal learner $q$ and the reference is the regret:
\begin{equation} \label{eq:regret}
R(z^N,x,y,q) = \log \frac{p_{\hat{\theta}(z^N,x,y)}(y|x)}{q_(y|z^N,x)}.
\end{equation}
As advocated in \cite{FogelFeder2018}, the chosen universal learner solves:
\begin{equation} \label{eq:minmax_prob}
\Gamma = R^*(z^N,x) = \min_q \max_y R(z^N,x,y,q)
\end{equation}
This min-max optimal solution, termed Predictive Normalized Maximum Likelihood (pNML), is obtained using ``equalizer'' reasoning, following \cite{shtar1987universal}:
\begin{equation}
q_{\mbox{\tiny{pNML}}}(y|z^N;x)=\frac{\pNMLSingle}{\sum_{y\in {\cal Y}} \pNMLSingle}.
\end{equation}
and its corresponding regret, independent of the true $y$, is:
\begin{equation} \label{eq:pNML_regret}
\Gamma = \log \left\{ \sum_{y\in {\cal Y}} \pNMLSingle \right\}.
\end{equation}

Note that this deviates from the commonly-used Empirical Risk Minimization (ERM) approach \citep{vapnik1992principles}, where in the log-loss case the learner chooses the model that assigns the maximal probability for the trainset:
\begin{equation} \label{eq:erm}
q_{\mbox{\tiny{ERM}}}(y|x) = \underset{p_\theta}{\textit{argmin }} \sum_{i=1}^{N}  -\log(p_\theta(y_i|x_i) .
\end{equation}
Nevertheless, it turns out that $\Gamma$ can also be used to obtain a bound on the performance of the ERM, see \cite{Fogel2018}.

The pNML has been derived for several model classes in related works \citep{FogelFeder2018} such as the barrier (or 1-D perceptron) model, and in \cite{BibasISIT} for the linear regression problem. In all these cases the advantages of the pNML and its corresponding learnability measure were discussed.

In some cases there are several possible model classes, or several possible algorithms. In this case, one can use a 'twice-universal' approach, see \cite{Fogel2018}, to achieve near-optimal performance not just within a class but over all possible classes.
In this approach, the best pNML learner from each of $K$ model classes is chosen $\{q_{k}\}_{k=1}^K$. 
Then, another pNML procedure is executed over all of these learners.

This paper's contribution is in proposing a scheme of the pNML learner for DNNs hypothesis class.
We show that the pNML improves upon the ERM learner performance in accuracy and log-loss on the testset, especially in the worst-case performance. 
For the first time, we show that the pNML is more robust to noisy labels in the trainset and adversarial attacks in the test. 
Furthermore, we show that $\Gamma$ may serve as a learnability measure for both the pNML and the ERM and that it can be used to point out when the trainset is composed of random labels and when the test sample is out of distribution or adversarial.
Finally, we demonstrate that the twice universal pNML scheme over the number of fine-tuned layers of DNNs forms has an even superior performance over both the pNML and the ERM.


\section{Related Work} \label{sec:related_work}
In this section, we briefly describe related work in universal prediction, model generalization, out of distribution detection and adversarial attack robustness.

\minisection{Universal Prediction and Online Learning}
In on-line learning prediction is done on a sequentially revealed sequence, where the loss is accumulated, and as the predicted label is revealed it essentially becomes part of the training.
This problem is an extension of the well-studied work on universal prediction, which is essentially online learning with no $x$'s.
As described in \cite{universal_prediction}, the universal prediction solution with log-loss provides a ``universal probability'' for the entire sequence, $q(y^N)$, which can be converted to a sequential prediction strategy via the chain rule $q(y^N)=\prod_{t=1}^N q(y_t|y^{t-1})$. 
The universal probability is given by either a Bayesian mixture $q(y^N) = \int_\Theta w(\theta) p_\theta (y^N)$ (stochastic setting) or by the normalized maximum likelihood (NML), $q_{\mbox{\tiny{NML}}}(y^N) = \frac{\max_{\theta} p_{\theta}(y^N)}{\sum_{\tilde{y}^N}\max_{\theta} p_{\theta}(\tilde{y}^N)}$, \citep{shtar1987universal}, (individual setting). Both solutions solve a corresponding minmax problem.
Online learning, with a feature sequence $x^N$ is less understood. 
Yet in \cite{Fogel2017} a universal learning solution is proposed that achieves vanishing regret in various settings.

\minisection{Model Generalization} 
Understating the model generalization capabilities is considered a fundamental problem in machine learning  \citep{vapnik2013nature}. This problem was mainly considered in the PAC learning framework, where several measures such as the VC-dimension and Rademacher complexity were suggested. However, those measures seems to fail in explaining the generalization of DNNs, see \cite{zhang2016understanding}. 
In this paper we advocate another measure, the pNML regret, which as demonstrated is valid also for DNNs.

\minisection{Out of distribution}
The problem of detecting out of distribution samples has been studied extensively for deep learning models.
It can be treated as classification problem where one tries to determine whether a test sample is from a different distribution than the training data.
Some works detect out of distribution test samples by post-processing the output of the model \citep{hendrycks2016baseline, liang2017enhancing} in order to generate uncertainty measure.
Others require a modification the neural networks or the training process \citep{andrews2016transfer}.
In \cite{gal2016dropout} it was suggested to use dropout as Bayesian inference approximations.
Again, in this paper we provide another technique to detect this situation with well established theoretical background.

\minisection{Adversarial Attack}
DNN models are vulnerable to adversarial perturbed samples that were designed to mislead a model at inference time.
A number of defensive techniques against adversarial attack in DNN have been proposed \citep{cisse2017parseval}. 
Adversarial training \citep{madry2017towards} appears to have the best results for learning robust models, however, it mostly protects against the attack it was trained for.
Unlike these methods, the method we present based on the pNML and its learnability measure, can be utilized for any learner, doesn't use adversarial samples in the training phase, not restricted to an architecture or the learning process and so it may improve any suggested procedure.

\section{Deep pNML} \label{sec:deep_pNML}
Given a trainset $z^N = \{(x_i,y_i) \}_{i=1}^N$, we assume the individual setting, where the data and labels of both the trainset and test sample are specific and individual values, i.e, there is no probabilistic connection between them.
As shown in \cite{FogelFeder2018}, the min-max optimal solution is known as the pNML
\begin{equation} \label{eq:pNML}
q_{\mbox{\tiny{pNML}}}(y|z^N;x)=\frac{\pNMLSingle}{\sum_{y\in {\cal Y}} \pNMLSingle}.
\end{equation}
Denote the normalization factor of the pNML as
\begin{equation}
C = \sum_{y\in {\cal Y}} \pNMLSingle,
\end{equation}
the distance between the pNML and a ''genie'' that knows the true test label is $\Gamma = \log C$.
Intuitively, to assign a probability for a possible outcome $q(y|z^N,x)$, the pNML adds the test sample $(x,y)$ to the trainset with a specific label, finds the best-suited model, and takes the probability it gives to that label. 
It then follows this procedure for every possible test label value. 
After iterating over all the possible labels of the test sample, an normalization is executed to get a valid probability assignment.
This method can be extended to any general learning procedure that generates a prediction based on a trainset. 
Such an algorithm can be the stochastic gradient descent (SGD) used to train neural networks.

Our proposed scheme of the pNML for DNNs is described in \myalgref{alg:deep_pNML} and consists of two steps: initial training and fine-tuning. First, we conduct the \textit{initial training} where we train a DNN using a given trainset with SGD. 
This is done using some given hyperparameters $\theta$, $\eta$, $\lambda$ and $steps$ which are the initial weights, learning rate, weight decay and the number of epochs respectively.
Next, we conduct \textit{fine-tuning}. 
Given a specific test example $x$, we examine every possible $y$. 
We add the pair $(x,y)$ to the trainset with a specific value for the test label, and perform several more SGD steps ($steps_2$ epochs) with the same hyperparameters. 
We save the prediction of the label we trained with.
We repeat this process for every possible value of the test sample.
The pNML predictor for a specific test label will simply be the predicted distribution for that label after the fine-tuning, normalized by the summation of all the corresponding probabilities to get a valid probability assignment.


For the rest of the paper, when referring to the ERM we will refer to the DNN whose weights were given by the initial training. The genie will be the DNN whose weights were obtained after the fine-tuning with the true test label. 
The pNML predictor will be the one obtained by taking the predictions for each label, each with the corresponding fine-tuned weights, and normalizing to get a valid probability function.

\begin{figure*}
\begin{minipage}[h]{0.49\columnwidth}
\null
\begin{algorithm}[H]
	\caption{Deep pNML}
	\label{alg:deep_pNML}
	\begin{algorithmic}
		\State {\bfseries Input:} $z^{N}$, $x$, $\eta$, $\lambda$, $\textit{steps}_1$, $\textit{steps}_2$
		\State Initialize $\hat{\theta}_0$ randomly
		\State Train: $\hat{\theta}_{\textit{ERM}} = SGD(\hat{\theta}_0, z^N, \eta, \lambda, \textit{steps}_1)$ 
		\For {i=1 {\bfseries to} $|\mathcal{Y}|$}
		\State $z^{N+1} = z^N \bigcup (x,y=i)$
		\State $\hat{\theta}_i = \textit{SGD}(\hat{\theta}_{\textit{ERM}}, z^{N+1}, \eta, \lambda, \textit{steps}_2)$  
		\State $p_i = p_{\hat{\theta}_i}(y=i|x)$
		\EndFor
		\State $C=\sum_{i=1}^{|\mathcal{Y}|}p_i$ 
		\State $\Gamma=\log{C}$
		\For{$i=1$ {\bfseries to} $|\mathcal{Y}|$} 
		\State $q_{\textit{pNML}}(y=i)= \frac{1}{C} p_i$
		\EndFor
		\State {\bfseries Return} $q_{\textit{pNML}}$,  $\Gamma$
	\end{algorithmic}
\end{algorithm}
\end{minipage}
\hfill
\begin{minipage}[h]{0.49\columnwidth}
\null
\begin{algorithm}[H]
   \caption{Twice Universality}
   \label{alg:twice_universality}
\begin{algorithmic}
   \State {\bfseries Input:} Test sample $x$, learners $\left\{q_k\right\}_{k=1}^K$
   \For {i=1  {\bfseries to} $|\mathcal{Y}|$}
   \State $p_i = \underset{k  \in \{1,..,K \}}{\textit{max }}q_{k}(y=i|x)$
   \EndFor
   \State $C^{\textit{TU}}=\sum_{i=1}^{|\mathcal{Y}|}P_i$ 
   \For{$i=1$ {\bfseries to} $|\mathcal{Y}|$} 
   \State $q_{\textit{pNML}}^{\textit{TU}}(y=i) =  \frac{1}{C^{\textit{TU}}} p_i$
   \EndFor
   \State {\bfseries Return} $q_{\textit{pNML}}^{\textit{TU}}$
   \State 
   \State 
   \State 
   \State 
   \vspace{0.31cm}
\end{algorithmic}
\end{algorithm}
\end{minipage}
\end{figure*}

\minisection{Twice universality} In many problems the best model class is unknown, and therefore being universal with respect to several model classes can improve the performance of the learner.
As part of our research, we implemented various versions of the pNML, where during the fine-tuning phase only some layers of the DNN's are updated. 
This can be seen as a nested hierarchy model class, where the richest model is the one where all the layers are updated in the fine-tuning phase, and the smallest class is that where none of them are updated and essentially the only model in the class is the ERM. 
We treat each of these variants as different classes $\{\Theta_k\}_{k=1}^K$, and evaluate their predictors, $q_k(\cdot|x)$, based on the pNML scheme in \myalgref{alg:deep_pNML}. 
Then, we executed the twice-universal algorithm described in \myalgref{alg:twice_universality} that produces the following learner: 
\begin{equation} \label{eq:twice_universal}
q^{\textit{TU}}_{\textit{pNML}}(y|z^{N},x)= 
\frac{\underset{k \in K}{\textit{max }} q_{k} (y|z^{N}, x, y)}{\sum_{y \in \mathcal{Y}}\underset{k \in K}{\textit{max }} q_{k} (y|z^{N}, x, y)}.
\end{equation}
Using the twice universal scheme, we can produce a min-max learner which has performance as good as the best learner from the various hypothesis classes.

\section{Applications} \label{sec:applications}
This section describes the applications of the pNML in deep neural networks \footnote{Code is available in \url{https://github.com/kobybibas/deep_pnml_experiments}}. 
First, we show how the pNML can improve upon the ERM learner in both the log-loss and accuracy sense.  
Then, we examine a trainset where the labels are assigned randomly, and show, again, that the pNML outperforms the ERM, and that the amount of noise in the trainset can be detected using the pNML regret.
Next, we use the pNML regret measure $\Gamma$ for detecting out of distribution samples. 
We further demonstrate that the pNML learner that more robust to adversarial samples.
Finally, we execute the twice universality algorithm on the number of fine-tuned layers, and we show that it outperforms the best learner tuned to each class. 

\subsection{pNML prediction performance} \label{sec:pNML_vs_erm}
For our first experiment we have used ResNet-18 architecture \citep{he2016deep} and  CIFAR10 dataset \citep{krizhevsky2014cifar}. We executed initial training and fine-tuning as described in \myalgref{alg:deep_pNML}. The initial training consisted of 200 epochs using SGD with a batch size of 128 and a learning rate 0.1, with a decrease by 0.1 after 100 and 150 epochs. 
In addition, we used a momentum of 0.9, weight decay of $5 \cdot 10^{-4}$ and standard normalization and data augmentation techniques. 
During fine-tuning we allowed updates of the weights of the last residual block along with the last fully connected layer, a total of 37,504 trainable parameters. 
The fine-tuning consisted of 10 epochs with learning rate of 0.001.


\begin{table}[h]
\caption{\textbf{Performance comparison.} Using ResNet-18 on CIFAR10 testset.}
\label{table:nml_vs_erm}
\begin{center}
\begin{tabular}{lcccr}
\toprule
Method & Acc. & Loss mean & Loss STD \\
\midrule
ERM    &   0.9183 &  0.194 & 0.822 \\
PNML   &   0.9184 &  0.167 & 0.314 \\
genie  &   0.9875 &  0.027 & 0.231 \\
\bottomrule
\end{tabular}
\end{center}
\end{table}

\begin{figure}
\centering
\begin{subfigure}[h]{0.49\linewidth}
\includegraphics[width=1.0\linewidth]{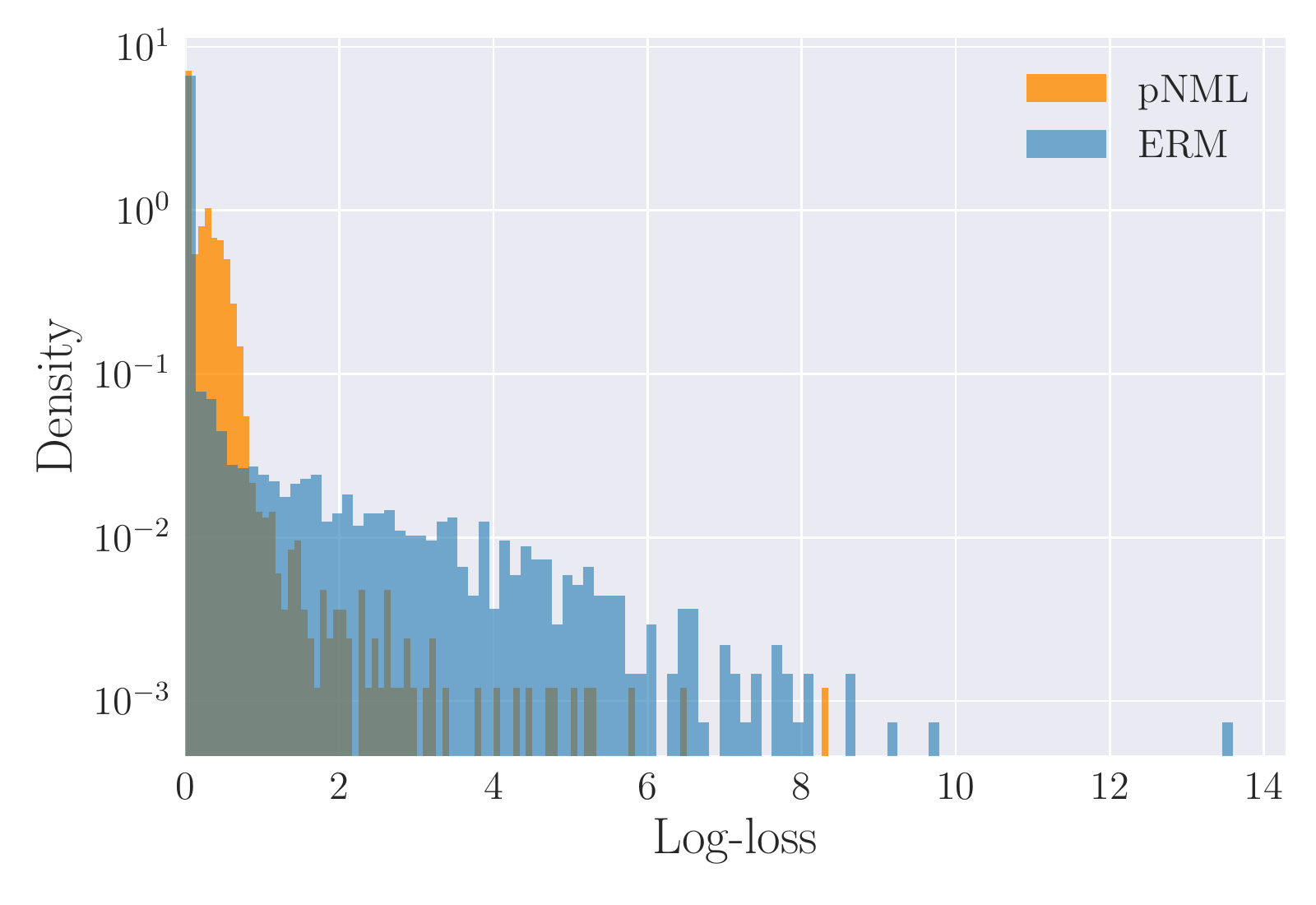}
\caption{Log-loss histogram for pNML and ERM. \label{fig:logloss_hist}}
\end{subfigure}
\begin{subfigure}[h]{0.49\linewidth}
\includegraphics[width=1.0\columnwidth]{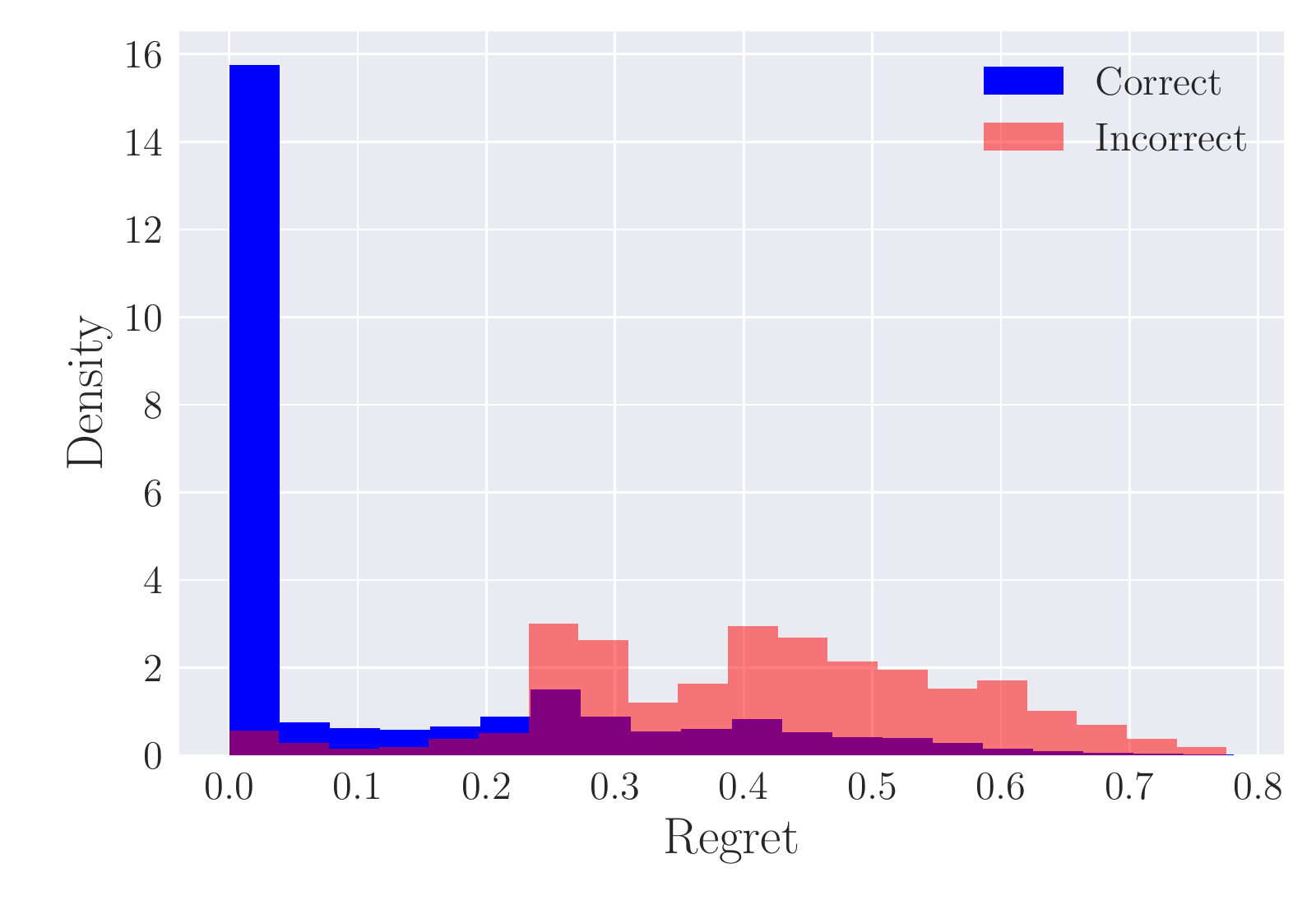}
\caption{pNML Regret histogram. \label{fig:correct_incorrect_hist}}
\end{subfigure}
\caption{\textbf{Log-loss and regret histograms.} (a) shows the log-loss for CIFAR10 testset with ResNet-18 architecture. The histogram is plotted for both pNML and ERM with density in semi-log scale. 
The log-loss is in base 10. 
(b) presents the regret histogram with correctly classified and miss-classified separation. 
The histogram is plotted for the pNML leaner with ResNet-18 architecture, trained and tested on CIFAR10 dataset.}
\end{figure}

The comparison between the ERM, pNML and the genie for the 10,000 test samples of CIFAR10 dataset is summarized in \tableref{table:nml_vs_erm}. 
We can see a $15\%$ improvement in the log-loss with a slightly better accuracy rate. 
More notable is the $60\%$ improvement in the standard deviation of the log-loss, which suggests that the pNML manages to avoid large losses. This property of the pNML can also be observed by the log-loss histogram of the ERM and the pNML, presented in \figref{fig:logloss_hist}.

\subsection{The regret as a confidence measure}
\begin{figure}
\centering
\begin{subfigure}[h]{0.49\linewidth}
\includegraphics[width=1.0\columnwidth]{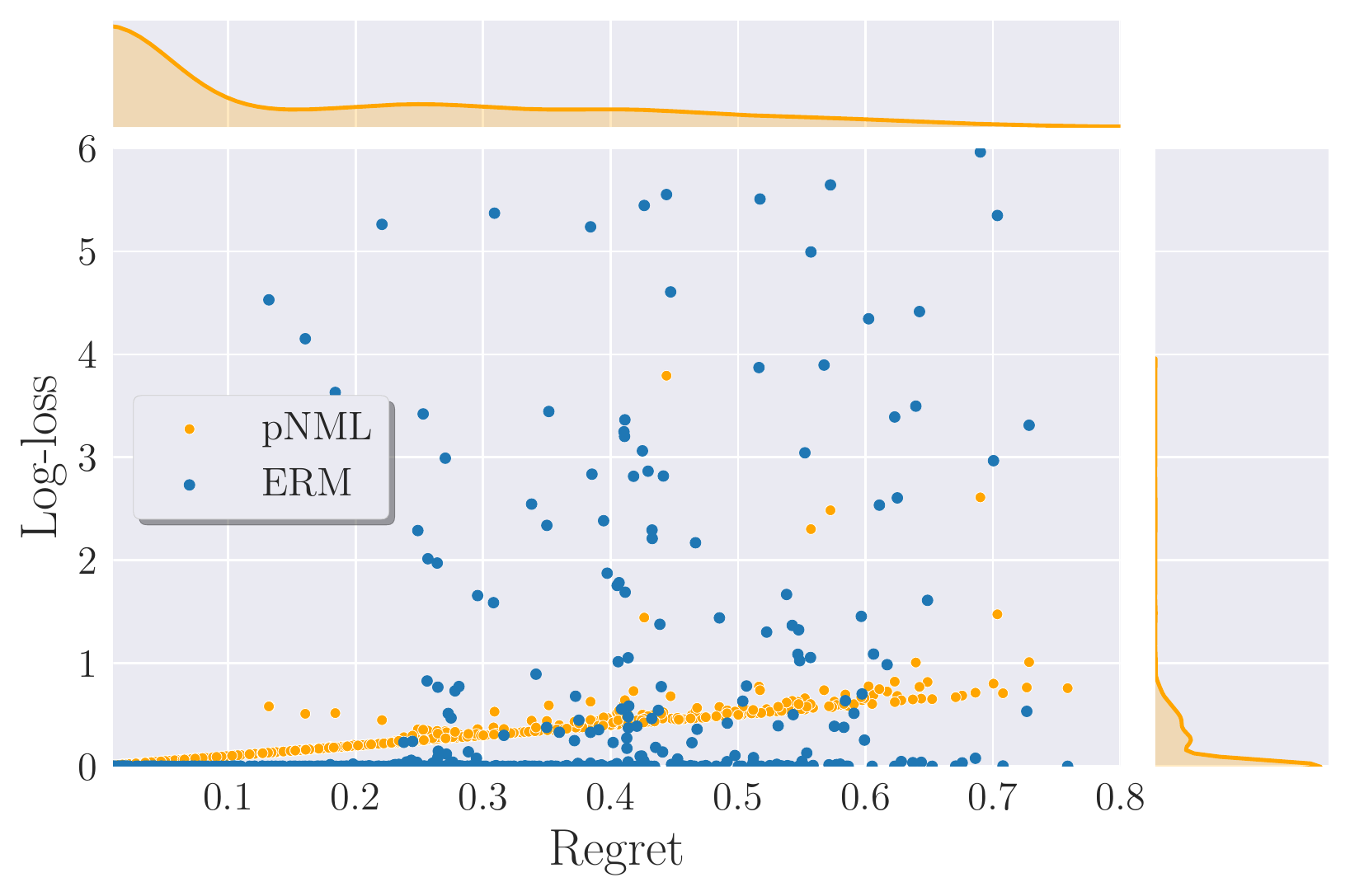}
\caption{Scatter plot of the log-loss and the regret. \label{fig:loss_vs_regret_scatter}}
\end{subfigure}
\begin{subfigure}[h]{0.49\linewidth}
\includegraphics[width=1.0\columnwidth]{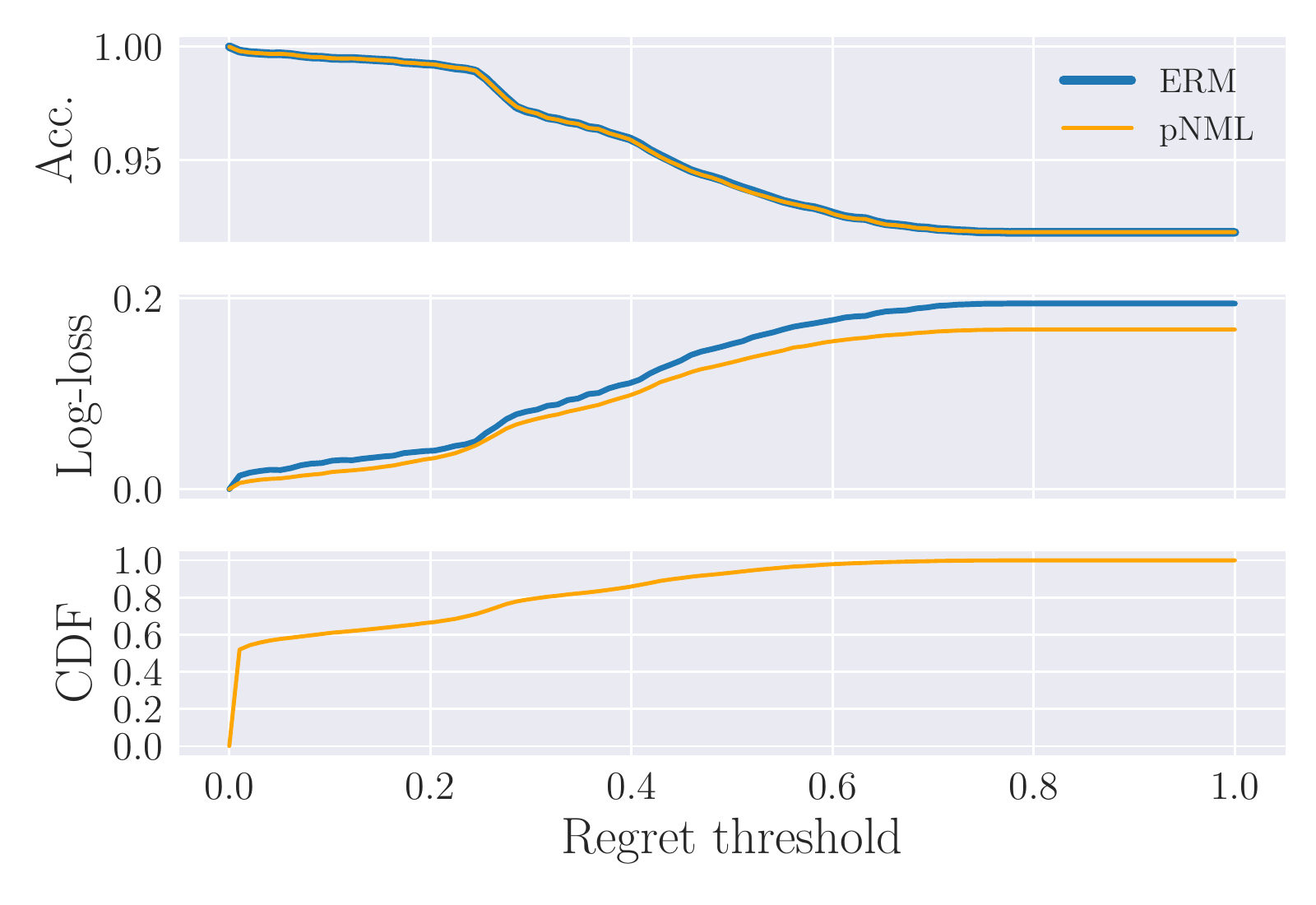}
\caption{Regret based classifier performance. \label{fig:regret_based_classifier}}
\end{subfigure}
\caption{\textbf{CIFAR10 testset regret.} In the bottom left of (a) a scatter plot of the loss and the regret of the pNML and the ERM are shown. In (a) figure's top and (a) figure's right the marginal distributions in a semi-log scale of the pNML regret and loss are presented. 
(b) shows the performance on samples with regret lower than a threshold. 
In the top, middle and bottom the accuracy, loss and Cumulative Distribution Function (CDF) of the testset are shown respectively.}
\end{figure}

\figref{fig:correct_incorrect_hist} shows the histogram of the regret with a distinction between correctly and incorrectly predicted samples. Clearly, this figure shows how the regret histograms are distinct and how the histogram for correctly classified samples indicates lower typical regret values than the corresponding values of incorrectly classified. Another interesting plot is given in \figref{fig:loss_vs_regret_scatter}. This figure shows a scatter plot of the log-loss of both the pNML and ERM as a function of the pNML regret of the test samples, along with the empirical marginal distributions of the loss and the regret in semi-log scale.  
One can notice that for large regret there is a tendency for high log-loss of the ERM, and in these cases, for the same sample the ERM loss is higher (sometimes much higher) than that of the pNML. 
One can also observe the almost straight line representing the pNML loss as a function of $\Gamma$.
This is because the pNML loss is $\Gamma$ plus the genie loss, and so $\Gamma$ may be interpreted as an `insurance'' the pNML pays to protect against large loss. 

These results imply that the pNML regret ($\Gamma$) is indeed a valid confidence measure for both the pNML and the ERM. 
To deepen our understanding of that conclusion, we evaluated the performances of both predictors when taking into account samples whose regret is not larger than some threshold, as shown in \figref{fig:regret_based_classifier}. 
When predicting about 70\% of the test samples, whose regret (measured at $\log$ base $10$) is less than $0.23$, the pNML predictor correctly classifies 99\% of them with log-loss smaller than about $0.041$.  
The ERM's results are similar, indicating that the regret is also a good confidence measure for the ERM learner.

\subsection{Random labels}  \label{sec:random_labels}
Recent evidence shows that it is possible to train DNNs on data with randomly generated labels and still get a perfect accuracy rate over the trainset \citep{zhang2016understanding}. 
This may question the generalization property of DNNs, as clearly such a model cannot generalize despite its perfect fit to the data.
Interestingly, $\Gamma$ which measures the distance from the genie can be used to reflect the generalization capabilities for a specific trainset.

In order to examine this situation, we trained a WideResNet-18 model \citep{zagoruyko2016wide} on CIFAR10 dataset, where some samples from the trainset were assigned random labels. 
We performed the initial training using stochastic gradient descent with a learning rate of 0.01 for 350 epochs without any regularization nor data augmentation. 
We ensured that the model accuracy rate on the trainset is 1.0. 
The fine-tuning phase consisted of 6 epochs with a learning rate of 0.01, changing the weights of only the last residual block along with the last fully connected layer.
We repeated the experiment with a variety of probabilities of the train samples labels to be random: 0.0, 0.2, 0.4, 0.6, 0.8 and 1.0.

\begin{table}[h]
\caption{\textbf{Random labels performance.} Performance on CIFAR10 testset when the trainset labels have the following probabilities to be random: 0.0, 0.2, 0.4, 0.6, 0.8, and 1.0.}
\label{table:random_prob}
\begin{center}
\begin{tabular}{lcccccc}
\toprule
Property & 0.0  & 0.2 & 0.4 & 0.6 & 0.8 & 1.0 \\
\toprule
ERM acc.  & 0.84 &    0.72 &    0.46 &    0.31 &    0.15 &    0.09\\ 
PNML acc. & 0.85 &    0.74 &    0.47 &    0.30 &    0.17 &    0.09\\
\cmidrule{1-7}
ERM loss  & 0.51 &    1.00 &    2.63 &    3.67 &    5.49 &    6.29\\
PNML loss & 0.49 &    0.83 &    0.90 &    0.97 &    1.12 &    1.28\\
\cmidrule{1-7}
Regret    & 0.49 &	  0.81 & 	0.86 &	  0.91 & 	0.90 & 	  0.89\\
\bottomrule
\end{tabular}
\end{center}
\end{table}

The results of this experiment are presented in \tableref{table:random_prob}.
Note that the regret is considerably larger when a significant part of the trainset has random labels, allowing us to detect when the model was trained in such a manner.
In addition, it is evident that the pNML is more robust to noisy training data in the log-loss sense than the ERM, while there is no significant difference in the accuracy.

\subsection{Detecting out of distribution examples} \label{sec:out_of_distribution}

Another situation where one can use a confidence measure is detecting out of distribution examples. Theoretically, a good learner or predictor should assign a low confidence to his predictions regarding out of distribution examples. 
Nevertheless, DNNs frequently produces high-confidence predictions, arguably because softmax probabilities are computed with the fast-growing exponential function \citep{hendrycks2016baseline}. 
Thus, it would be interesting to see if $\Gamma$ is indeed large for out of distribution examples.

\begin{figure}
\centering
\begin{subfigure}[h]{0.49\linewidth}
\includegraphics[width=1.0\linewidth]{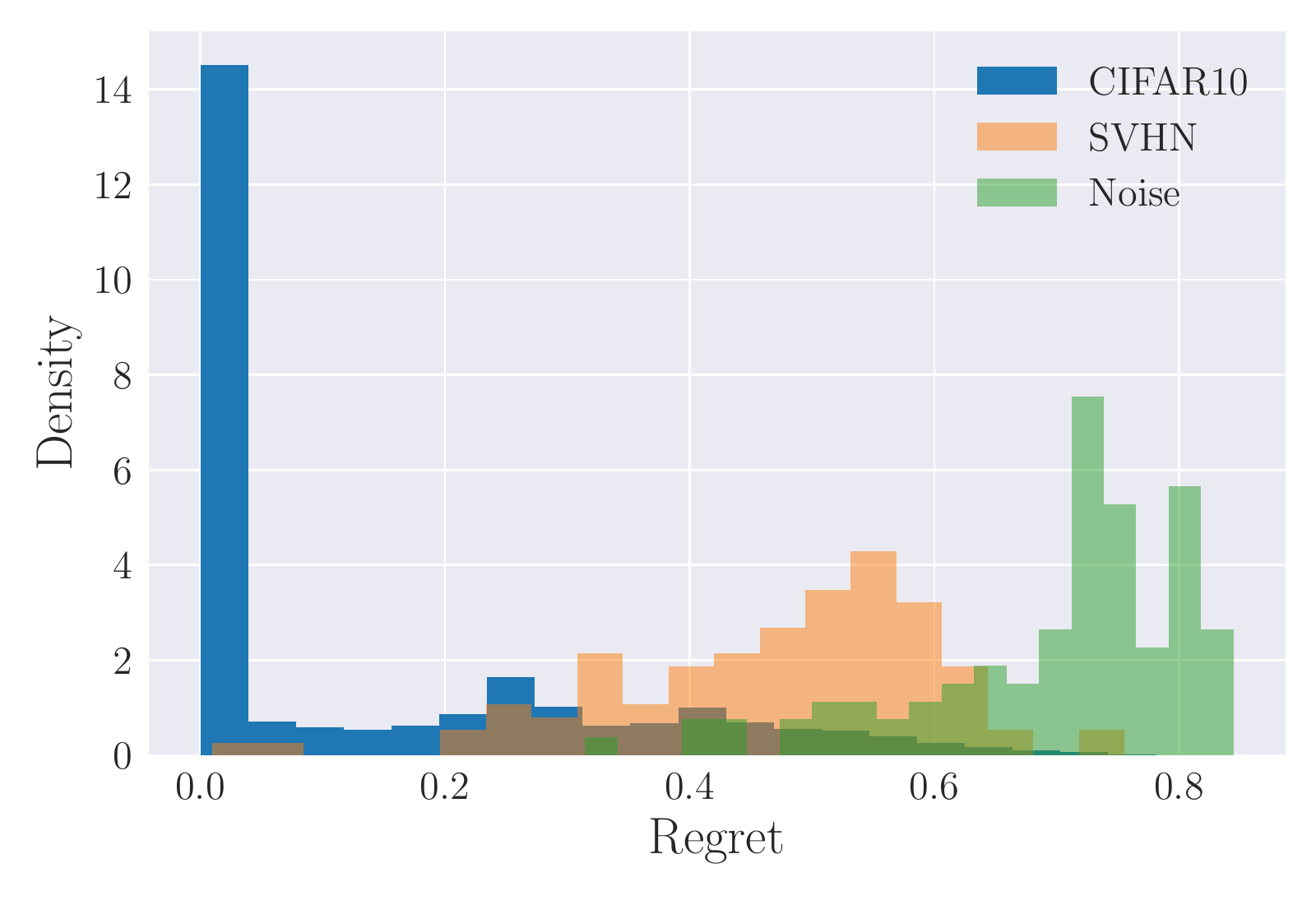}
\caption{Out of distribution regret histogram. \label{fig:out_of_dist_hist}}
\end{subfigure}
\begin{subfigure}[h]{0.49\linewidth}
\includegraphics[width=1.0\columnwidth]{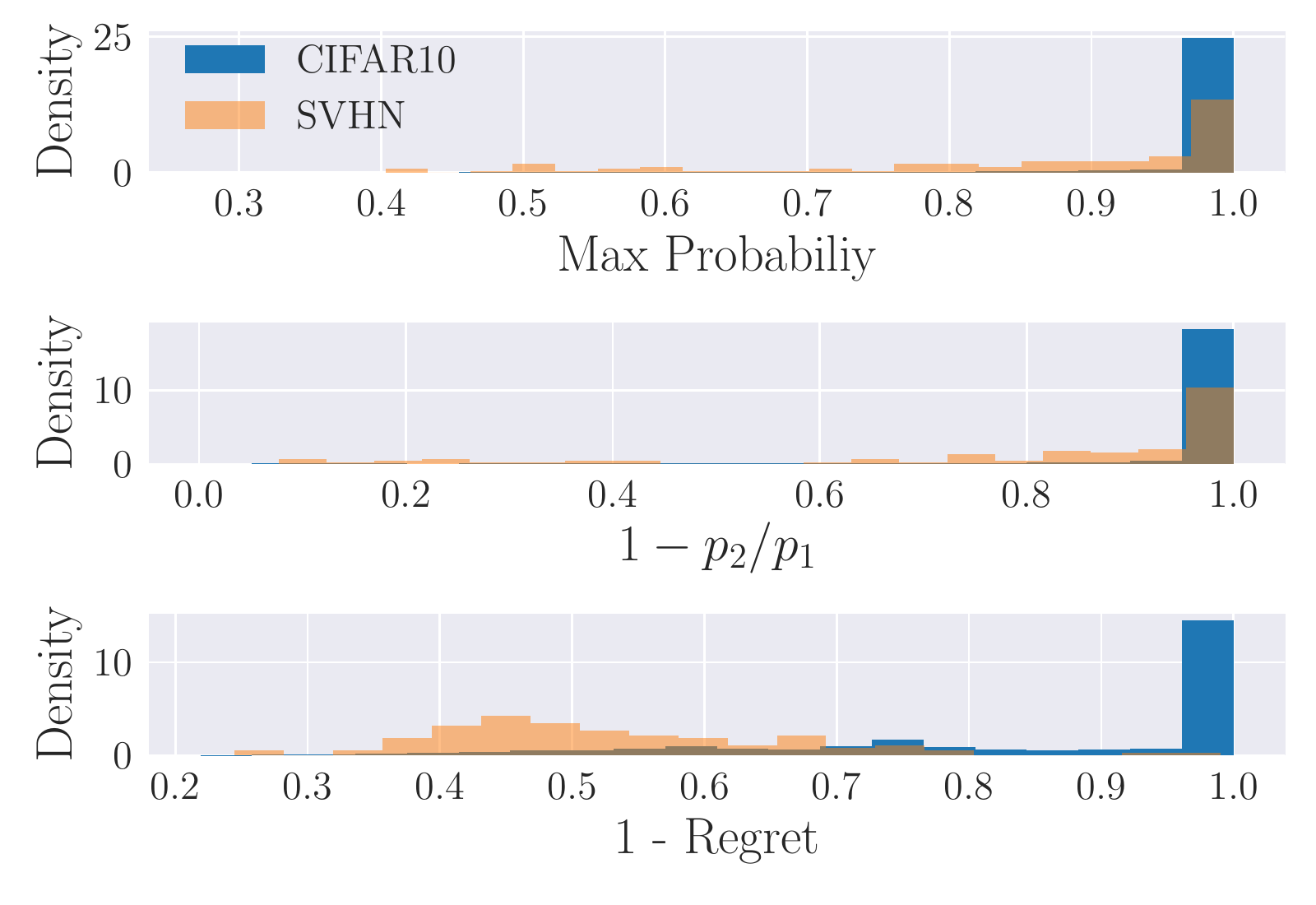}
\caption{In and Out of distribution histograms. \label{fig:ood_detection_svhn}}
\end{subfigure}
\caption{\textbf{Out of distribution histograms.} Resnet-18 trained on CIFAR10 trainset. 
(a) shows the regret histograms of test samples from CIFAR10, SVHN and random noise images. (b) presents the empirical distribution of various uncertainty measures on SVHN test samples.}
\end{figure}

\begin{table}
\caption{\textbf{Out of distribution detection.} 
Comparison between the different criteria for noise and SVHN images. 
The metrics are mean KL divergence and log-likelihood ratio test. The higher the value of metrics, better detection of out of distribution samples.}
\label{table:out_of_dist}
\begin{center}
\begin{tabular}{lr}
\begin{tabular}{lccc}
\toprule
Testset                 & Model      & $D_{\textit{KL}}$   & $D_{\textit{LRT}}$ \\
\midrule
\multirow{ 3}{*}{Noise} & Max prob   & 0.794  & 0.113 \\
\cmidrule{2-4}
                        & $1-p_2$/$p_1$ & 0.643  & 0.164\\
\cmidrule{2-4}
                        & Regret     & 3.583  & 0.658 \\
\bottomrule
\end{tabular}
&
\begin{tabular}{lccc}
\toprule
Testset                 & Model      & $D_{\textit{KL}}$   & $D_{\textit{LRT}}$ \\
\midrule
\multirow{ 3}{*}{SVHN}  & Max prob   & 0.777  & 0.108 \\
\cmidrule{2-4}
                        & $1-p_2$/$p_1$ & 0.701  & 0.153 \\
\cmidrule{2-4}
                        & Regret     & 2.950 & 0.535 \\
\bottomrule
\end{tabular}
\end{tabular}
\end{center}
\end{table}

To check that, we performed the initial training with the ResNet-18 architecture and CIFAR10 dataset as in Sec.~\ref{sec:pNML_vs_erm}.
We then performed the fine-tuning phase, again in a similar manner to that described in Sec.~\ref{sec:pNML_vs_erm}, examining 10,000 test samples from CIFAR10, 100 test samples from SVHN \citep{netzer2011reading} and 100 random Gaussian noise images. 
The regret histograms for the different testset are presented in \figref{fig:out_of_dist_hist}, showing a clear distinction between in distribution and out of distribution samples. 

To further evaluate the performance of the regret as a way to detect out of distribution samples, we use common metrics that measure the differences between two probability distributions. Let $q_1$ denote the probability assignment of the in distribution samples and $q_2$ the probability assignment of the out of distribution samples. 
The first metric to measure the difference between the probabilities is the average Kullback Leibler (KL) divergence:
\begin{equation}
D_{\textit{KL}}(q_1, q_2) = \frac{1}{2}\left(KL(q_1||q_2) + KL(q_2||q_1)\right)
\end{equation}
The second metric comes from the likelihood ratio test (LRT), see \cite{cover2012elements}.
Denote the probability assignment
\begin{equation}
q_\lambda = \frac{q_1^\lambda(x) q_2 ^{1-\lambda}(x)}{\sum_{x \in \mathcal{X}} q_1^\lambda(x) q_2 ^{1-\lambda}(x)} 
\end{equation}
and let $\lambda^*$ be the point for which the KL divergence between $q_1$ and $q_\lambda$ equals to the KL divergence between $q_2$ and $q_\lambda$. The value of the KL divergence at this point is the distance
\begin{equation}
D_{\textit{LRT}}(q_1, q_2)=D(q_{\lambda^*}||q_1) = D(q_{\lambda^*}||q_2)
\end{equation}

We compare our method to the maximum probability method which determines if the test sample is from the in distribution or out of distribution based on the maximum of probability assignment \citep{hendrycks2016baseline} and to $1-p_2/p_1$, where $p_1$ is the maximum probability of the prediction and $p_2$ is the second largest value of the probability assignment \citep{mor2018confidence}.

The comparison is shown in \tableref{table:out_of_dist}. 
One can note that by all the metrics we considered, the regret is indeed a significantly better classifier of out of distribution samples than the max probability and $1-p_2/p_1$ uncertainty measures for both noise images and SVHN images.
Figure \ref{fig:ood_detection_svhn} (Top) shows the histogram of in distribution and out of distribution test images based on the max probability of the ERM output and Figure \ref{fig:ood_detection_svhn} (Middle) shows the histogram based on $1-p_2/p_1$ criterion. 
The histogram of in distribution and out of distribution test images based on the pNML regret is shown in figure \ref{fig:ood_detection_svhn} (Bottom). 
When categorizing samples according to the regret there is a significant separation between the two distributions, whereas when using the max probability criterion or the $1-p_2/p_1$ criterion the two histograms overlap almost entirely.

\subsection{Adversarial Attack} \label{sec:adversarial_attack}
We evaluated the performance of our approach for adversarial samples. Our goal is in both detecting adversarial samples and devising a robust learner against them.
We created the adversarial samples using the Fast Gradient Sign Method (FGSM) \citep{goodfellowexplaining}: First, we computed the sign of the loss function gradients according to the input and then multiplied these signs by a small value $\epsilon$ to create a perturbation.
The adversarial image is the original image after the addition of the perturbation: 
\begin{equation}
x  \leftarrow x + \epsilon \textit{ sign}(\nabla _x \mathcal{L}(\theta,x,y)).  \end{equation}
We generated adversarial samples on CIFAR10 dataset using a different model than the one used for testing (black box setting), namely ResNet-32 with $\epsilon$ equal 0, 0.001, 0.005, 0.01 and 0.05 (the pixels values are between -1.0 and 1.0) for 500 test samples. 

We implemented the pNML with the same parameters as in Sec.~\ref{sec:pNML_vs_erm} and with the ResNet-18 architecture, and tested on adversarial samples.
The performances of the pNML, ERM and the genie for each $\epsilon$ are presented in Table \ref{table:adversarial}.
The pNML and the ERM have similar accuracy rates.
However, the log-loss of the pNML is much better for every $\epsilon$ and especially for $\epsilon$ of 0.05, where the pNML log-loss is 1.72 compared with the ERM  log-loss of 3.22.
It is interesting to note that the genie, which is trained with the true test label, also has a significant decrease in accuracy, with only 53\% success for $\epsilon$ that equals to 0.05.
Table \ref{table:adversarial} shows that when the perturbation strength is increased, the mean regret is also increased. This behavior implies that the adversarial image might be detected based on the regret. 

The evaluation of the adversarial images detection using the regret is shown in Table \ref{table:adversarial_detection} with the same metrics as in Sec.~\ref{sec:out_of_distribution}. 
It shows that there is some difference in the regret between regular and adversarial samples, which is increased when $\epsilon$ is increased. 
Nevertheless, the difference is much more subtle than that obtained in the out of distribution detection case, even for large values of $\epsilon$.

\begin{table}
\centering
\caption{\textbf{Adversarial attack performance and detection.} Learners performances on adversarial samples from CIFAR10 testset. The adversarial samples were generated with FGSM Method in black box settings. $\epsilon$ indicates the strength of the attack. 
$D_{KL}$ and $D_{LRT}$ indicate effectiveness of the attack detection (see Sec.~\ref{sec:out_of_distribution}).}
\label{table:adversarial} \label{table:adversarial_detection}
\begin{tabular}{lr}
\begin{tabular}{lccccc}
\toprule
Property \textbackslash \ $\epsilon$ & 0.0  & 0.001 & 0.005 & 0.01 & 0.05\\
\toprule
ERM acc.  &     0.92 &    0.88 &    0.77 &    0.68 &    0.26 \\ 
pNML acc. &     0.92 &    0.87 &    0.77 &    0.69 &    0.27 \\
Genie acc. &    0.99 &    0.96 &    0.92 &    0.86 &    0.53 \\
\cmidrule{1-6}
ERM loss  &     0.19 &    0.31 &    0.70 &    1.11 &    3.22 \\
PNML loss &       0.18 &   0.20 &   0.35 &    0.52 &    1.72 \\
Genie loss &    0.03 &    0.07 &    0.17 &    0.31 &    1.39 \\
\cmidrule{1-6}
Regret    &     0.15 &    0.13 &    0.18 &    0.22 &    0.33 \\
\bottomrule
\end{tabular}
&
\begin{tabular}{lcc}
\toprule
$\epsilon$ \textbackslash \ Metric & $D_{\textit{KL}}$  & $D_{\textit{LRT}}$ \\
\toprule
0.001   &     0.03   &    0.01    \\
0.005   &     0.05   &    0.01    \\
0.01    &     0.14   &    0.03    \\
0.05    &     0.67   &    0.17    \\
\bottomrule
\end{tabular}
\end{tabular}
\end{table}

\subsection{Twice Universality}
In addition to the universality inside a model class $\Theta$, a second universality can be made regarding the index of multiple model families $\left\{\Theta_k\right\}_{k=1}^{K}$. This is the `twice universal' approach. We consider three model classes: The first class is the one described in Sec.~\ref{sec:pNML_vs_erm} where the fine-tuning affected only the last two layers. In the second class fine-tuning affected all layers. The third class is simply the DNN obtained after the initial training, which is the ERM.

Our next results were obtained by twice-universality over these 3 classes: For each test sample and each label, we took the maximum of the pNML probabilities over the 3 model classes. Then we normalized the probability of the test sample in order to create a valid probability assignment, as described in \myalgref{alg:twice_universality}.
\tableref{table:twice_universality} summarizes the results on CIFAR10. The twice universal predictor is better than any other class predictor in the log-loss sense. 
In addition, its accuracy is 0.92 which is as good as the 2 layer model. 

\begin{table}
\caption{\textbf{Twice Universality performance.} The performance of the pNML learners with different amount of layers that were finetuned.}
\label{table:twice_universality}
\begin{center}
\begin{tabular}{lr}
\begin{tabular}{lccc}
\toprule
Dataset & Model & Acc.  & Loss\\
\hline
\multirow{4}{*}{CIFAR10}&   0 layers    & 0.920 &  0.203 \\
                        &   2 layers    & 0.921 &  0.173 \\ 
                        &   7 layers    & 0.913 &  0.244 \\
                        &   Twice Univ. & 0.920 &  0.168 \\
\bottomrule
\end{tabular}
&
\begin{tabular}{lccc}
\toprule
Dataset & Model & Acc.  & Loss\\
\hline
\multirow{4}{*}{MNIST}  & 0 layers      & 0.930 &   0.093 \\
                        & 1 layer      & 0.937  &   0.086 \\ 
                        & 2 layers      & 0.937 &   0.085 \\
                        & Twice Univ.   & 0.950 &   0.081 \\
\bottomrule
\end{tabular}
\end{tabular}
\end{center}
\end{table}

We also examined the twice universality method on MNIST dataset \citep{lecun-mnisthandwrittendigit-2010}. 
The network used was a simple multilayer perceptron \citep{orhan2011eeg}, with 10 hidden units. 
The initial training consisted of 100 epochs with a learning rate of 0.01 for the first 10 epochs, 0.01 for the next 30, 0.001 for the next 40 and 0.0001 for the rest. 
Next, we executed the pNML procedure of \myalgref{alg:deep_pNML}. 
The fine-tuning phase consisted of 10 epochs with a learning rate of 0.0001, changing either the two layers or only the last one.
The results of the twice universality method on MNIST dataset are also presented in \tableref{table:twice_universality}. 
The twice universal learner has the best performance, both in the log-loss and the accuracy rate. Since this learner is close to the optimal one for every specific sample, it seems that it avoids the pitfalls of each of the other learners and thus manages to outperform them.  

\section{Conclusion} \label{sec:conclusion}
In this paper, we presented the Deep pNML leaner in the individual setting with respect to the log-loss function. 
We showed that the pNML scheme outperforms the ERM in the accuracy and log-loss sense, it is more robust for noisy trainset and has better performance for adversarial samples.
In addition, the inherent regret property of the pNML can detect out of distribution and adversarial samples as well as noisy trainsets.
One could assume that when executing the pNML scheme, the DNN would fit exactly to each possible test label and therefore using the regret as a generalization measure might be useless. 
We show that in a ``learnable'' situation this is not the case and that the regret is an informative generalization measure.
Specifically, we showed that the pNML pointwise generalization measure ($\Gamma$) is valid for DNNs, where other generalization measures (like the VC dimension) fail.
Finally, we introduce the twice universality concept and showed its performance on CIFAR10 and MNIST dataset with different DNN architectures, making it universal on any hyper-parameter choice, yet still at least as good as the best one.

As for future work, there is on-going work on the theory of the pNML and its generalizations, including relating the pNML procedure to stability analysis. In the more practical aspects related to DNNs, it is interesting to explore the effect of using a varying number of epochs in the fine-tuning phase. 
This may enhance our understanding of DNNs and perhaps improve performance. 
In addition, we plan to try the pNML for various network architectures and hyper-parameters, with the hope to improve performance and to get more insight on how they affect the generalization abilities of the network. 

\vskip 0.2in
\bibliography{main_deep_pnml}

\end{document}